\documentclass[journal]{IEEEtran}

\usepackage{cite}
\usepackage[pdftex]{graphicx}
\graphicspath{{figures/}}
\DeclareGraphicsExtensions{.jpg}
\usepackage{authblk}
\usepackage{amsmath}
\usepackage{array}
\usepackage{caption}
\usepackage{xcolor}
\usepackage[caption=false,font=footnotesize]{subfig}
\usepackage{url}
\usepackage{cleveref}

\begin{document}

\title{An Iterative Labeling Method for Annotating Fisheries Imagery}

\author[1]{Zhiyong~Zhang}
\author[1]{Pushyami~Kaveti}
\author[1]{Hanumant~Singh}
\author[2]{Abigail~Powell}
\author[3]{Erica~Fruh}
\author[3]{M.~Elizabeth~Clarke}
\affil[1]{Field Robotics Laboratory, Northeastern University}
\affil[2]{Lynker Technologies, NWFSC, NOAA}
\affil[3]{Northwest Fisheries Science Center, NOAA}

\markboth{Journal of \LaTeX\ Class Files,~Vol.~14, No.~8, August~2021}%
{Shell \MakeLowercase{\textit{et al.}}: Bare Demo of IEEEtran.cls for IEEE Journals}

\maketitle

\begin{abstract}
In this paper, we present a methodology for fisheries-related data that allows us to converge on a labeled image dataset by iterating over the dataset with multiple training and production loops that can exploit crowdsourcing interfaces. We present our algorithm and its results on two separate sets of image data collected using the Seabed autonomous underwater vehicle. The first dataset comprises of 2,026 completely unlabeled images, while the second consists of 21,968 images that were point annotated by experts. Our results indicate that training with a small subset and iterating on that to build a larger set of labeled data allows us to converge to a fully annotated dataset with a small number of iterations. Even in the case of a dataset labeled by experts, a single iteration of the methodology improves the labels by discovering additional complicated examples of labels associated with fish that overlap, are very small, or obscured by the contrast limitations associated with underwater imagery.
\end{abstract}


\IEEEpeerreviewmaketitle

\section{Introduction} 
\IEEEPARstart{T}{echnologies} for imaging the deep seafloor have evolved significantly over the last three decades 
\cite{Perspectives_in_visual_imaging_for_marine_biology_and_ecology_from_acquisition_to_understanding}. Manned submersibles, Remotely Operated Vehicles, Autonomous Underwater Vehicles \cite{SeaBED_AUV_offers_new_platform_for_high_resolution_imaging}, towed vehicles \cite{taylor2008evolution}, and bottom mounted and mid water cameras \cite{amin2017modular} have all contributed to an explosion in terms of our ability to obtain high resolution, true color \cite{kaeli2011improving} camera imagery underwater.  

These technologies should ultimately enable us to conduct more efficient fishery independent surveys yielding improved stock assessment and ecosystem based management \cite{francis2007ten}. Efforts are underway to analyze the imagery manually, and with various levels of automation based on tools from machine learning for a variety of fisheries and habitat monitoring applications including coral reefs \cite{Use_of_machine_learning_algorithms_for_the_automated_detection_of_cold_water_coral_habitats_A_pilot_study}, \cite{Automated_classification_of_underwater_multispectral_imagery_for_coral_reef_monitoring} \cite{Imaging_coral_I_imaging_coral_habitats_with_the_SeaBED_AUV}, starfish \cite{Toward_Robust_Image_Detection_of_Crown_of_Thorns_Starfish_for_Autonomous_Population_Monitoring}, seastars, \cite{Automated_Counting_of_the_Northern_Pacific_Sea_Star_in_the_Derwent_Using_Shape_Recognition}, scallops \cite{dawkins2017open} and commercially important groundfish \cite{Evaluating_the_SeaBED_AUV_for_Monitoring_Groundfish_in_Untrawlable_Habitat}. The reality, however, is that extracting actionable information from our large underwater image datasets is still a challenging task.

\begin{figure}[!t]
\centering
\includegraphics[width=3.2in]{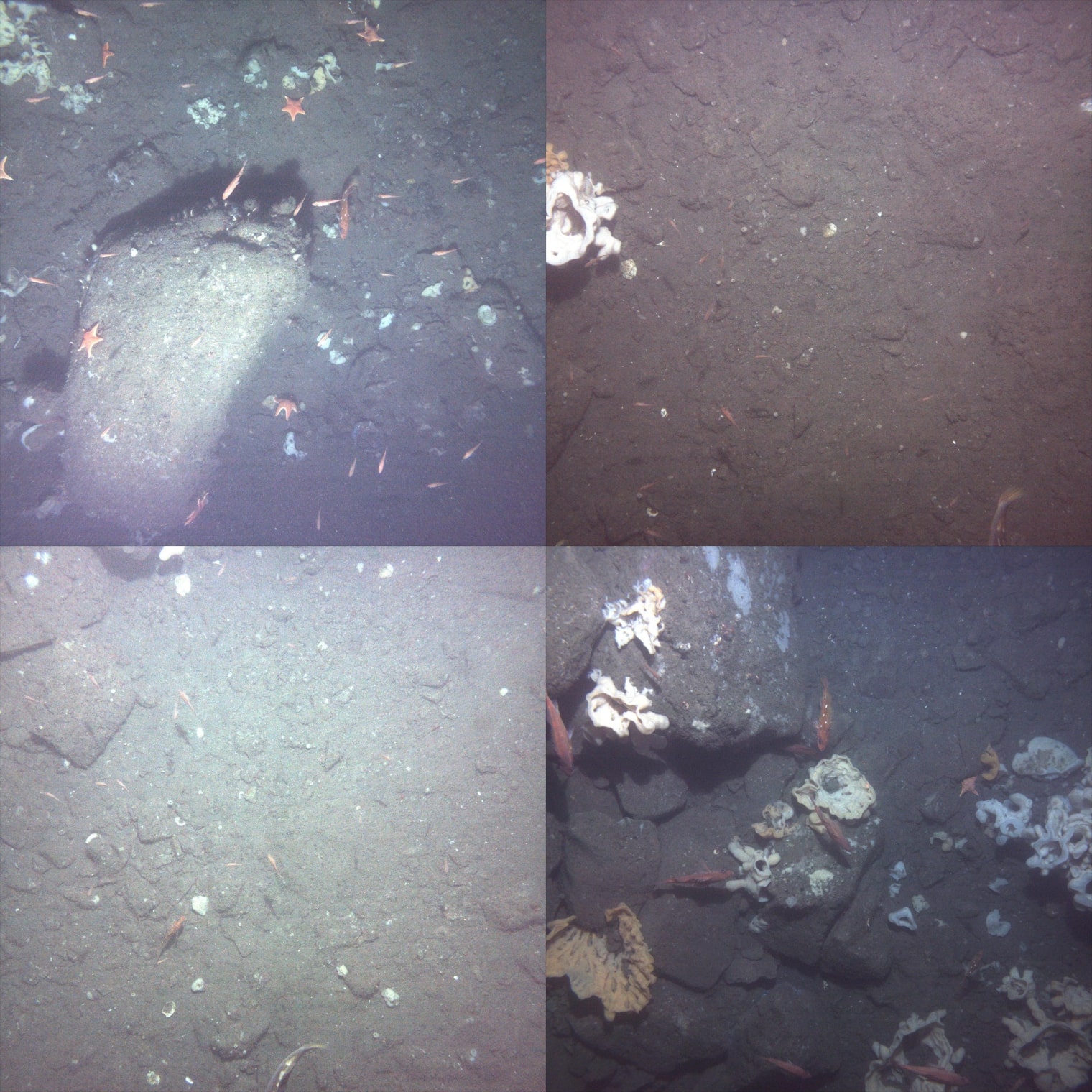}
\caption{Underwater image samples from one of the datasets with no annotations. There are very large fisheries related image datasets that are freely available but are not annotated. These would require significant efforts from experts in the field to label.}
\label{empty_examples}
\end{figure}

Machine learning techniques for land based applications have seen remarkable successes primarily due to the availability of large labeled datasets, such as ImageNet \cite{ImageNet_A_large_scale_hierarchical_image_database}, PASCAL VOC \cite{The_Pascal_Visual_Object_Classes_(VOC)_challenge}, and COCO \cite{Microsoft_COCO_Common_Objects_in_Context}, but the lack of similar large datasets related to fisheries is a primary factor limiting the widespread use of machine learning techniques.  Several efforts have been initiated for underwater imagery that include the ability for annotation \cite{VIAME}, \cite{A_Realistic_Fish_Habitat_Dataset_to_Evaluate_Algorithms_for_Underwater_Visual_Analysis}, \cite{Local_inter_session_variability_modelling_for_object_classification},
\cite{A_research_tool_for_long_term_and_continuous_analysis_of_fish_assemblage_in_coral_reefs_using_underwater_camera_footage}, \cite{Overview_of_the_Fish4Knowledge_Project}, \cite{LifeCLEF_2015_Multimedia_Life_Species_Identification_Challenges}, \cite{Wildfish_A_Comprehensive_Fish_Benchmark_for_Multimedia_Research}  \cite{richards_zooinverse} but the results, so far, have been limited. In fact, there are a large number of underwater image datasets available with no efficient means to label them. One such example is shown in (\cref{empty_examples}).

There are multiple options for labeling image data sets. The most simple  utilize crowdsourcing on a marketplace such as Mechanical Turk \cite{10.1007/978-3-642-35142-6_14}. Other options include professional annotation services and the use of domain experts. 

Mechanical Turk is fairly inexpensive but the results for specialized imagery, such as that associated with fisheries, are often mixed and unreliable. Our experience has shown that some workers annotate images with labels that are randomly placed, and requires a prohibitive amount of time and effort spent in approving or rejecting these results.
Professional annotation platforms are better than crowdsourcing with Mechanical Turk but at a cost that is almost an order of magnitude greater. The end results though better, are still not perfect. 
In sharp contrast to the other two methods, marine biologists and fisheries experts can also label the data set and these domain experts usually label exceptionally well. However, domain experts are few in number, very expensive, and often do not have the time to engage in such tasks. 

\begin{figure}[!t]
\centering
\includegraphics[width=3.2in]{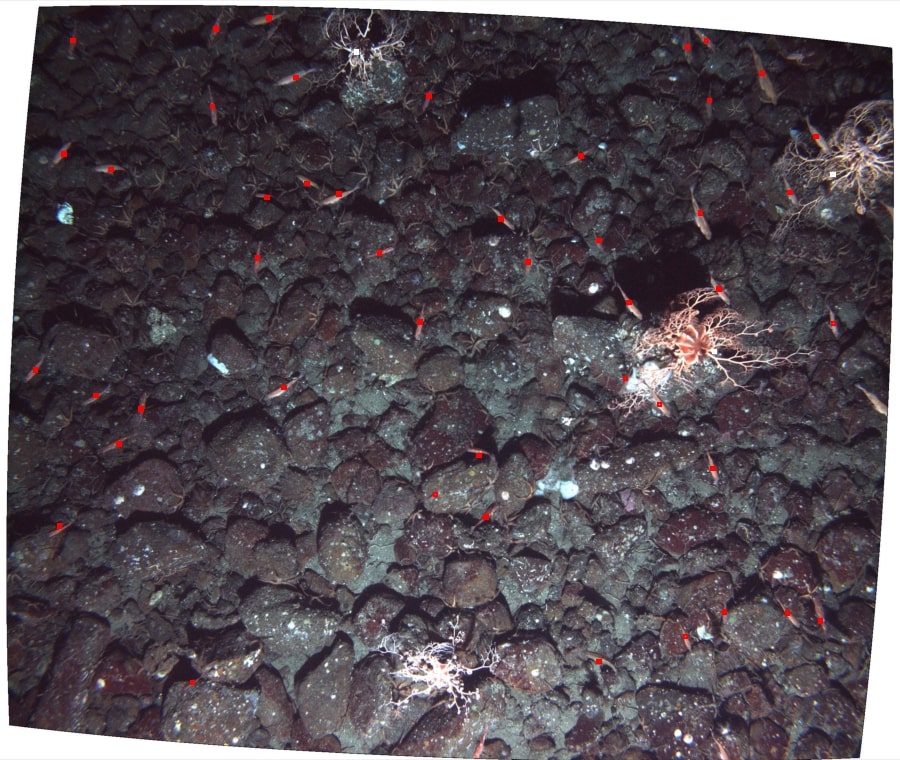}
\caption{Underwater images annotated by NOAA marine biologists, with dot annotations on each object. These annotations were not specifically carried out for Machine Learning which require bounding boxes as opposed to a dot to label bojects. We note that this image is undistorted and rectified for stereo matching.}
\label{NOAA_dot_annotation}
\end{figure}

Somewhere in the middle lie professional annotation services like Zooniverse\cite{zooni} which are far better than crowdsourcing, but at a cost that is often not economical for very large data sets. Label quality while better than crowdsourcing on a marketplace, can be variable as the annotaters are not experts in marine biology. They often miss objects in the images or confuse different species.

This work outlines a methodology that can label very large datasets, semi-automatically, by combining machine learning with Mechanical Turk crowdsourcing. We utilize a unique iterative process with an auto approval  that allows us to check the quality of the workers algorithmically, precisely and efficiently, without any human intervention. We can also use the same techniques for converting historical expert annotations (\cref{NOAA_dot_annotation}) to quickly create labeled data sets for machine learning that are critically required for fisheries and ecosystem based management applications.

\begin{figure*}[!t]
\centering
\includegraphics[width=7in]{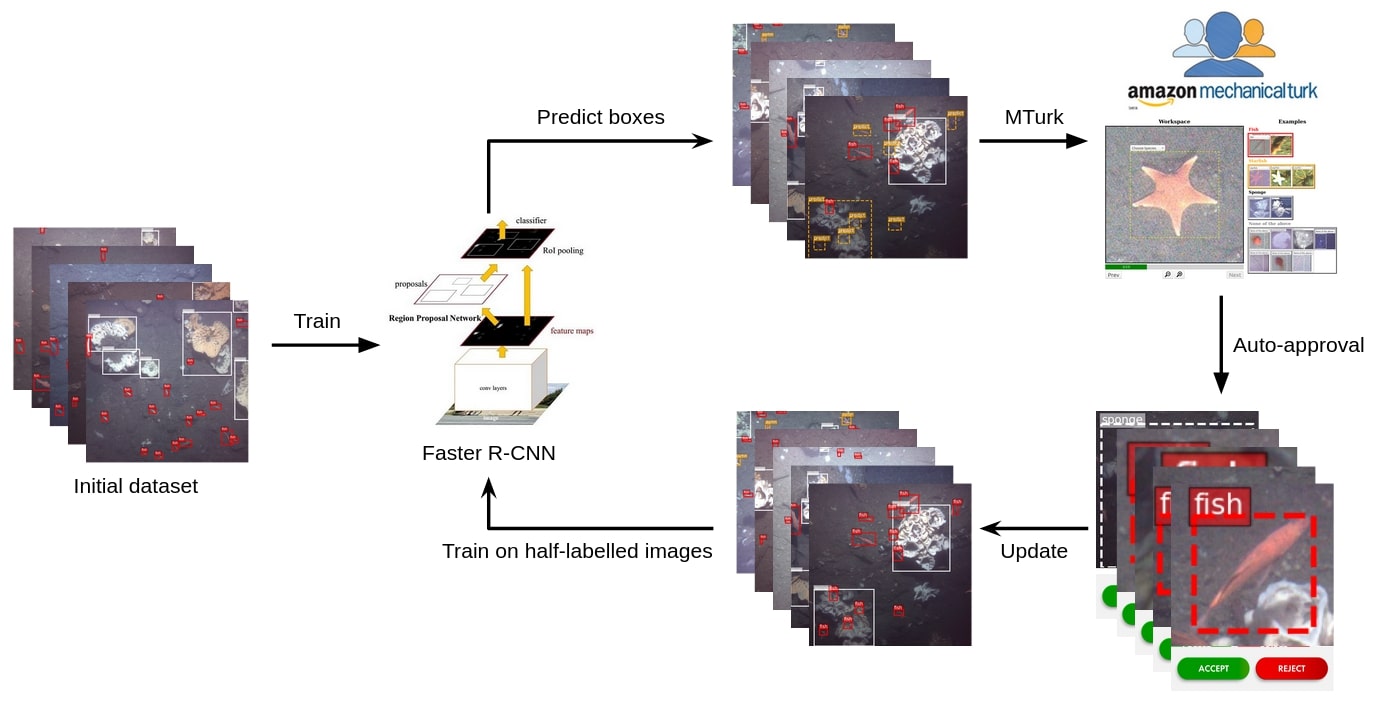}
\caption{The complete iterative labeling process: We train the initial dataset with Faster R-CNN \cite{Faster_R_CNN_Towards_Real_Time_Object_Detection_with_Region_Proposal_Networks} to get an initial model. We predict objects on the new image set with this initial model and publish the prediction boxes to MTurk for correction. A method for auto-approval filters the results. We then update the labels and train on half-labeled images to get a new model and prediction set for the next loop. We converge to a completely labeled dataset in 4-6 iterations.}
\label{diagram_1}
\end{figure*}

\section{Related work}

Deep convolutional neural networks are now ubiquitous for image segmentation \cite{Very_Deep_Convolutional_Networks_for_Large_Scale_Image_Recognition} and classification \cite{ImageNet_Classification_with_Deep_Convolutional_Neural_Networks}. 
Multiple works have explored the use of standard deep convolutional neural networks such as YOLO \cite{You_Only_Look_Once_Unified_Real_Time_Object_Detection}, DIGITS \cite{Digits_the_deep_learning_gpu_training_system}, and Tensorflow \cite{TensorFlow_A_system_for_large_scale_machine_learning} for fisheries related work including \cite{Towards_Automated_Fish_Detection_Using_Convolutional_Neural_Networks}, \cite{Fish_detection_and_identification_using_neural_networks_some_laboratory_results}, \cite{Vision_based_real_time_fish_detection_using_convolutional_neural_network}, \cite{Automated_Detection_of_Rockfish_in_Unconstrained_Underwater_Videos_Using_Haar_Cascades_and_a_New_Image_Dataset_Labeled_Fishes_in_the_Wild},  \cite{Deep_learning_for_benthic_fauna_identification}. However, None of these efforts address the task of generating large labeled datasets.

\subsection{Learning methods for dataset generation} 
As opposed to segmentation and classification, our focus is on high quality labeling of large underwater image datasets.  Iterative labeling methods can be divided into two classes, active and semi-supervised learning respectively. The former requires active human-in-the-loop participation while the latter does not. Most active learning methods rely on the generation of a high-performance detector, 
\cite{Large_scale_live_active_learning_Training_object_detectors_with_crawled_data_and_crowds}, \cite{Active_Learning_for_Deep_Object_Detection} but they do not guarantee high-quality annotations. For semi-supervised learning \cite{Iterative_Labeling_for_Semi_Supervised_Learning}, no human intervention is required, but the end result is not aimed at providing high-quality, labeled datasets.

Our approach is similar to active learning\cite{settles2009active} methods which fall under weakly supervised learning techniques. The goal of active learning is to learn from a small set of labels by choosing a selective and highly representative set of data samples. Our focus, however, is to create a superior, large, labeled dataset of multiple specialized categories corresponding to marine taxa. Hence, we query all the images as opposed to selecting a few samples. Our goal is to come up with a scalable, automated approach to minimize the effort in generating huge labeled datasets. In the following subsections we describe each component of our iterative labeling process in detail.

\subsection{Performance enhancement on crowdsourcing platforms}
Many human-machine collaboration methods have been proposed to improve the efficiency of human-in-the-loop annotation. \cite{Visual_Recognition_with_Humans_in_the_Loop} presents an interactive, hybrid human-computer method
for image classification. Deng et al. \cite{Scalable_multi_label_annotation} work on multi-label annotation, which find the correlation between the objects in the real world to reduce human's computation time of checking the existence in the image. Russakovsky et al. \cite{Best_of_both_worlds_Human_machine_collaboration_for_object_annotation} ask human annotators to answer a series of questions to check and update the predicted bounding boxes while Wah et al \cite{Multiclass_recognition_and_part_localization_with_humans_in_the_loop} query the user for binary questions to locate the part of the object. Vijayanarasimhan et al. \cite{Multi_Level_Active_Prediction_of_Useful_Image_Annotations_for_Recognition} incrementally update the classifier by requesting multi-level annotations, from full segmentation to a present/absent flag on the image.

Our work is most similar to LSUN \cite{LSUN_Construction_of_a_Large_scale_Image_Dataset_using_Deep_Learning_with_Humans_in_the_Loop}. LSUN also hides the true ground-truthed labels in the MTurk task to verify worker performance and enable auto-approval. It also uses two workers to label the same image for quality control. In contrast to LSUN, We only label once during the iterative labeling process, and we define our task as working with individual objects in an image as opposed to considering all the objects in an entire image.

Other works related to our efforts, focus on taking advantage of Amazon MTurk with enhanced interfaces or other mechanisms. Kaufmann et al. \cite{More_than_fun_and_money_Worker_Motivation_in_Crowdsourcing_A_Study_on_Mechanical_Turk} adapt different models from classic motivation theory, work motivation theory and test the effect of extrinsic and intrinsic motivation. \cite{The_relationship_between_motivation_monetary_compensation_and_data_quality_among_US_and_India_based_workers_on_Mechanical_Turk} study the relationship between motivation, monetary compensation, and data quality among US and India-based workers on MTurk. Spatharioti et al.  \cite{On_Variety_Complexity_and_Engagement_in_Crowdsourced_Disaster_Response_Tasks} studied the effects of on switching subtask type at different frequencies to impact the measures of worker engagement. Kaveti et al. \cite{Role_of_intrinsic_motivation_in_user_interface_design_to_enhance_worker_performance_in_Amazon_MTurk} design an MTurk interface and evaluate based on self-determination theory and add a guided practice test to achieve higher annotation accuracy. Bhattacharjee et al. \cite{Process_Design_to_Use_Amazon_MTurk_for_Cognitively_Complex_Tasks} simplify complex tasks on MTurk by combining batches, dummy variables, and worker qualification. Our work in this regard is simpler as we do away with qualification tests and tutorials, to lower the barriers for workers to enter our tasks. We can get away with these simplifications as we have already provided the simplest form of task to the Mechanical Turk workers.

\section{Iterative labeling process}
\begin{figure}[!t]
\centering
\subfloat[loop 1 predict]{\includegraphics[width=1.6in]{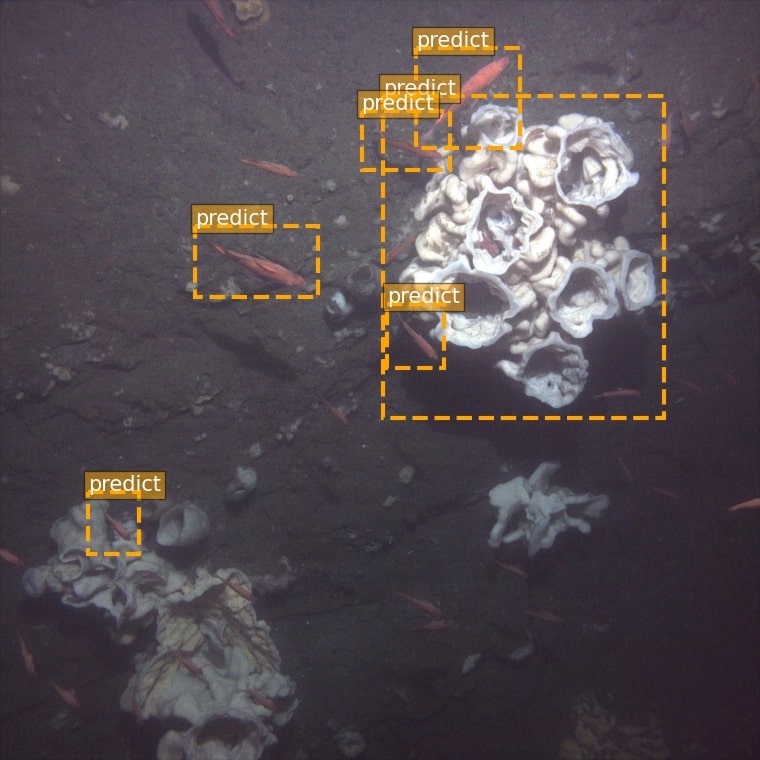}}
\hfil
\subfloat[loop 1 update]{\includegraphics[width=1.6in]{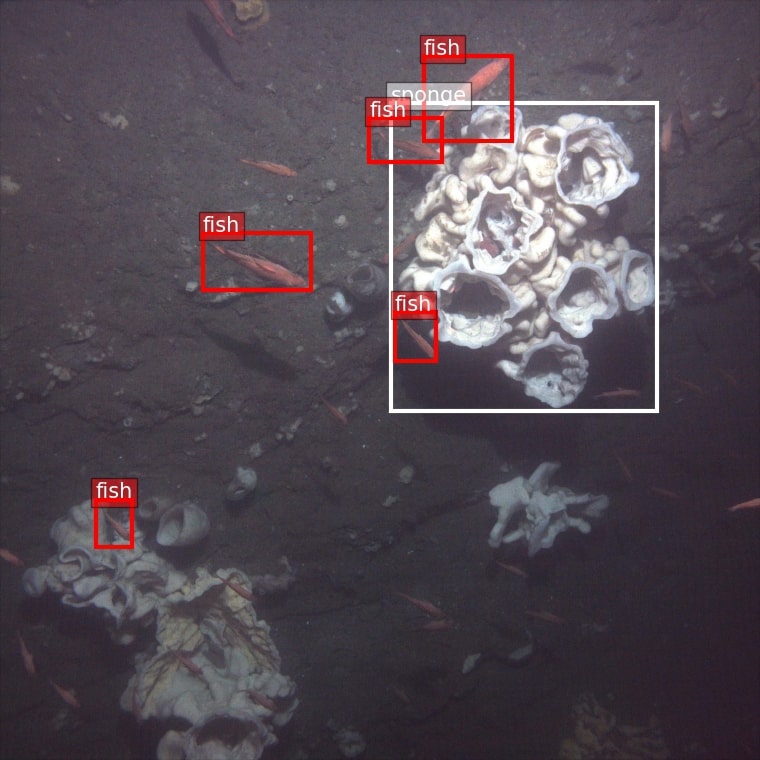}}
\hfil
\subfloat[loop 2 predict]{\includegraphics[width=1.6in]{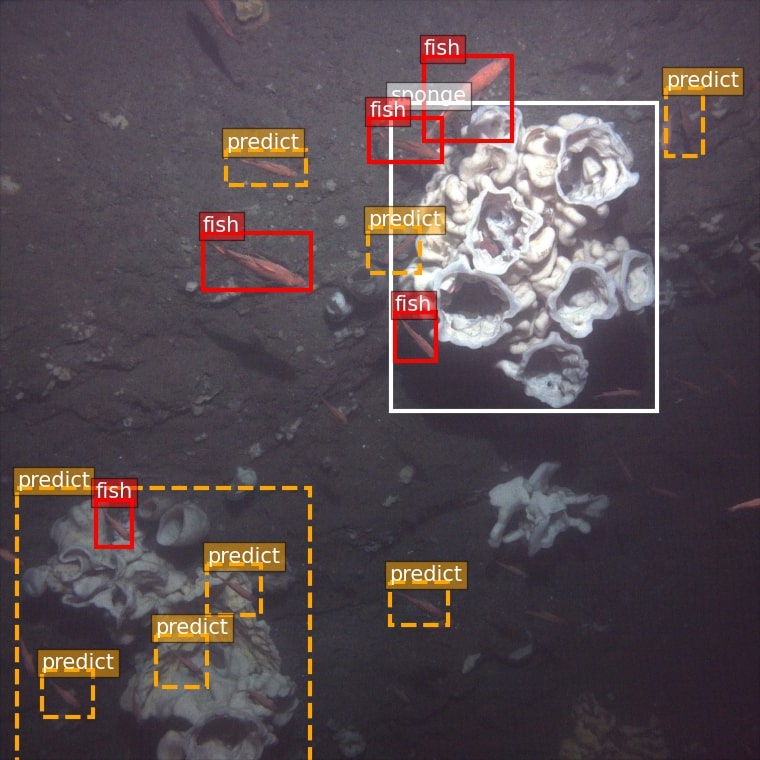}}
\hfil
\subfloat[loop 2 update]{\includegraphics[width=1.6in]{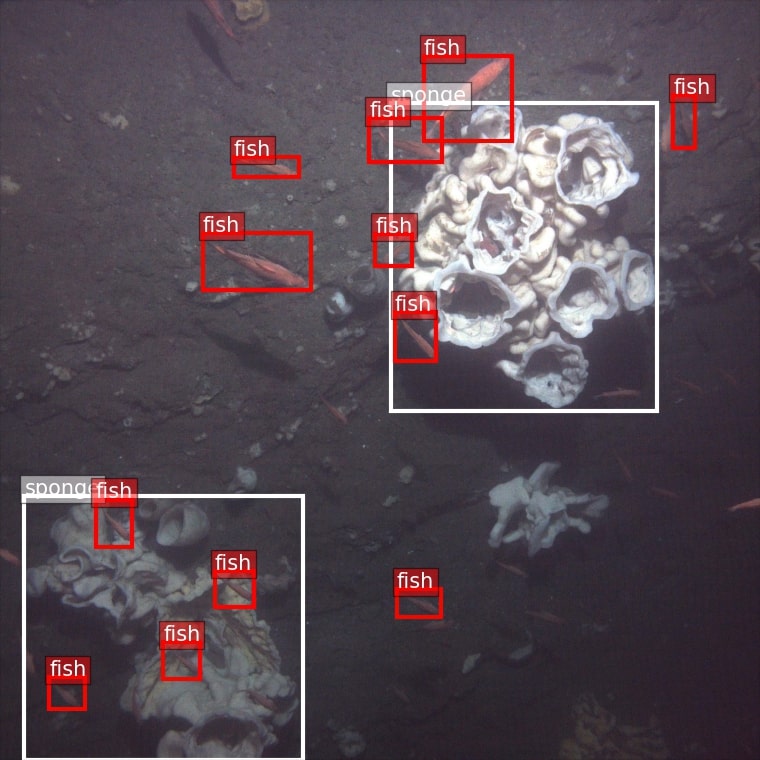}}
\hfil
\subfloat[loop 3 predict]{\includegraphics[width=1.6in]{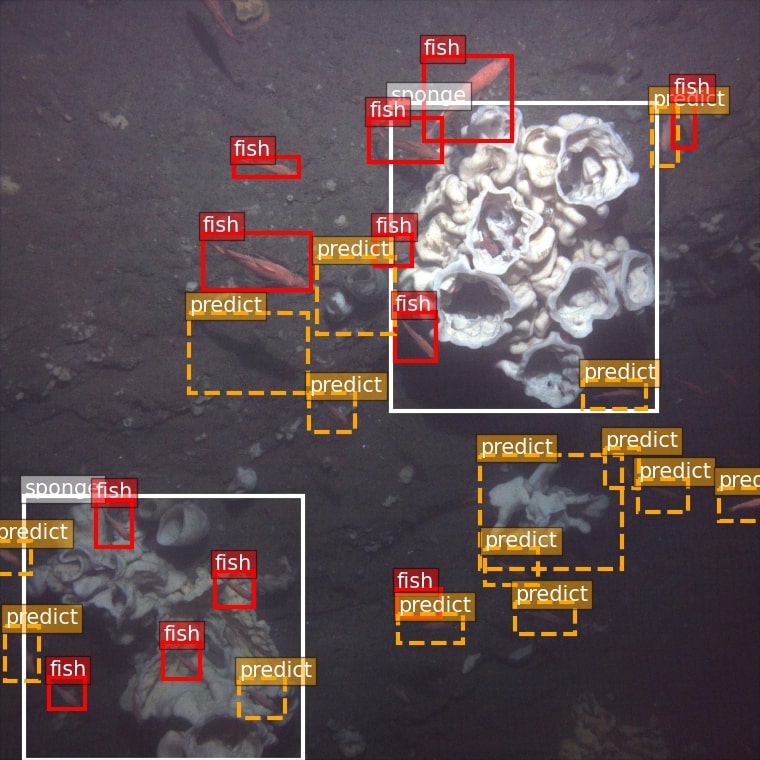}}
\hfil
\subfloat[loop 3 update]{\includegraphics[width=1.6in]{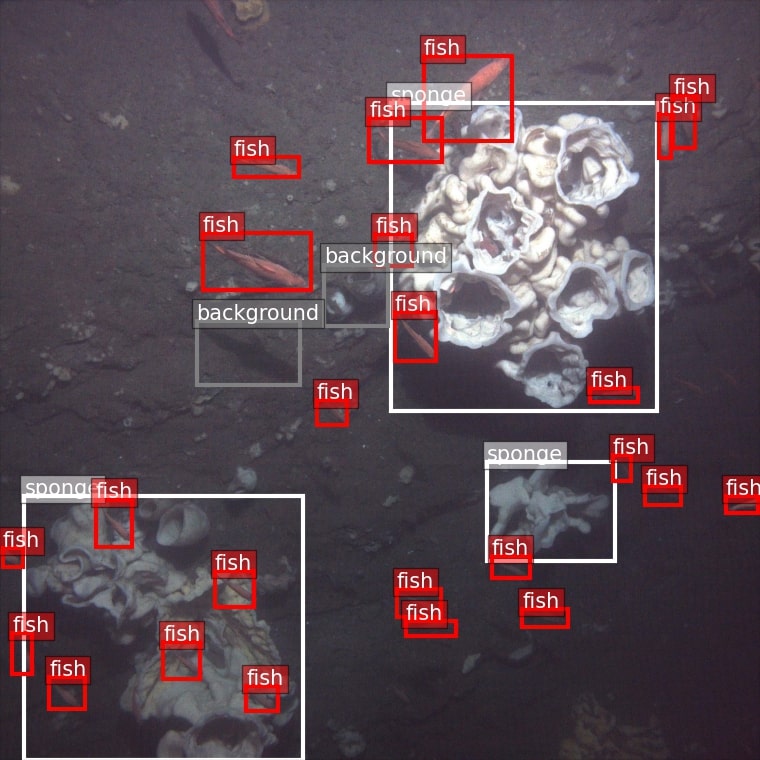}}
\caption{An example of the iterative labeling process. Orange dashed boxes are the predictions of each loop. These prediction boxes are published to MTurk for correction. The updated labels based on the MTurk results are then used for the next loop.}
\label{predict_update_loop}
\end{figure}

The overview of our method for the iterated labeling process for underwater images is ilustrated in \cref{diagram_1}. The process starts with building an initial deep learning model for making bounding box predictions on a small subset collection of underwater images. These predictions   are published to a crowd-sourcing platform with a well-designed assistive interface for validation. An auto-approval method filters bad labels from the crowdsourcing platform. The filtered labels are added to the dataset and used for further training to get new predictions. Thus, we start with a small set of annotations and increase the number of annotations every loop until all objects in all images have been labeled. Fig. \ref{predict_update_loop} shows an example of the predict-update loop for a single image.

\subsection{Initial Model}
We start with a small seed dataset labeled by fisheries experts. This serves as our initial dataset which we use to train our deep learning object detection model.  The seed dataset consists of three categories of marine species labeled namely Rockish, Starfish and Sponge with 965 instances of Rockfish, 650 instances of Starfish and 2005 instances of Sponge. This data is not large enough to completely train the model but as long as the annotations of the seed dataset contain a large number of examples of objects of all the categories, the initial model is sufficient to make reasonable predictions to feed into the first iteration.
As the iterative labeling process does not have real-time constraints, we chose Faster R-CNN as the object detection network. We also chose ResNet-101 \cite{Deep_Residual_Learning_for_Image_Recognition} as the backbone network as the image dimensions (2448 x 2050), require a deep network architecture. We built the network based on the Detectron2 \cite{Detectron2}. We train the object detection network on 2 RTX 2080 GPUS with a batch size of 2 for 60 epochs. We obtain a mean Average Precision(mAP) of 0.79 on our validation dataset consisting of 300 images.

Once we have the initial model trained, we use it to make predictions of the learned object categories on a new unlabeled image dataset. Since there is no groundtruth for this data, we do not know if these predictions are true positives. We leverage the workers within Mechanical Turk to classify and correct the predictions.  
\begin{table}[!t]
\renewcommand{\arraystretch}{1.3}
\caption{initial dataset}
\label{initial_dataset}
\centering
\begin{tabular}{ccc}
\hline
Rockfish & Starfish & Sponge\\
\hline
965 & 650 & 2005\\
\hline
\end{tabular}
\end{table}

\subsection{Assistive Annotation Interface Design}

In this section we describe the design and development of the user interface on MTurk used to facilitate the human in the loop learning process. One of the key aspects of the interface is to present the user with a convenient way to determine if the predictions made by the deep learning model are accurate and, if so, to annotate them. The correct object detections are used as groundtruth labels to continue the training process of the deep learning model. The main idea is that over a series of these predict-correct-update-train loops we will end up with a superior model.

The most common interface design for labeling object instances in images on MTurk requires the workers to detect all the objects in the image and draw bounding boxes for each object before moving on to the next image. This process can be cumbersome when there are a lot ($> 30$) of instances per image to label, and is especially difficult when the dataset consists of unique specialized categories of objects. This also affects the motivation of the worker to perform the task as explained in our previous work \cite{Role_of_intrinsic_motivation_in_user_interface_design_to_enhance_worker_performance_in_Amazon_MTurk}. We make a few novel design choices in constructing the MTurk annotation interface as described below. A snapshot of our assistive annotation interface is shown in \cref{assistive_interface}

\subsubsection{Tutorial/examples of annotations} One of the challenges of underwater datasets is that they contain unique and uncommon objects and the workers on MTurk come from diverse backgrounds with variability in experience and expertise \cite{}. We dedicate a small portion of the interface to showcase a set of labeling examples for the various marine species encountered in the dataset. This helps to familiarize workers with the dataset.

\subsubsection{Labeling cues} Instead of asking the workers to find all possible instances of the categories on a raw image we provide several labeling cues to make it easy for the workers. We show the predictions made by the deep learning model as a dashed bounding box. The workers are then asked to adjust it to tightly fit the object and choose the species for it via a dropdown menu. These features help to correct localization and classification losses during supervision. Sometimes, the background in the images can be mistakenly predicted as a species so a "None of the above" option (corresponding to the background) was added to the species drop-down menu. 

\subsubsection{UI controls}
The images in our underwater dataset can contain 40-50 instances of relevant objects per image. Sometimes these instances tend to be really small and are occluded by other objects due to overlap. We can see this in \cref{predict_update_loop}. Thus we choose to zoom in and show each bounding box prediction as opposed to showing all the boxes at the same time. In this way, workers can focus on one single object instead of the whole image. This is beneficial for labeling tiny objects and also amplifies user performance while adjusting the bounding boxes.

\begin{figure}[!t]
\centering
\includegraphics[width=3.2in]{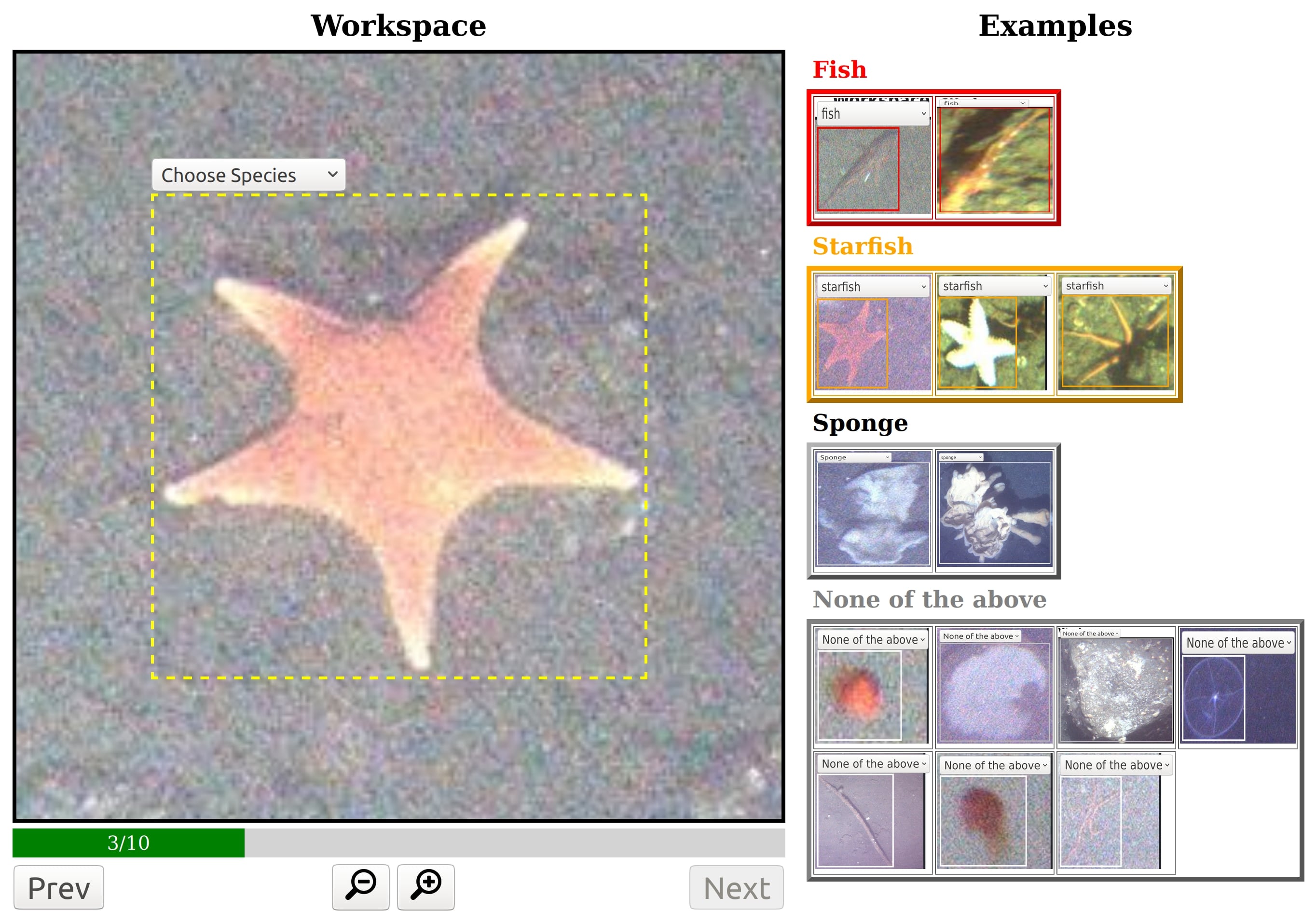}
\caption{The assistive interface for MTurk workers. Workers can focus on the object we provide. Simply identifying the species and fitting the bounding box dramatically improves the reliability of the labels. Ground-truth is hidden at the last task to implement auto-approval.}
\label{assistive_interface}
\end{figure}

\subsection{Auto-Approval}
The biggest drawback of the Mturk platform is with respect to the quality control of the workers. Workers might have minimal to zero knowledge of the domain and usually have low attention spans \cite{paolacci2014inside}. MTurk does allow one to select workers based on some criteria or via a test but, the requesters often end up spending a lot of time and resources reviewing annotation results. This negates the purpose of wanting to create a fully automated human-in-the-loop annotation process. Thus we have developed an auto-approval mechanism to to assess how well workers are doing, and accept or reject the annotations without any intervention.

We put ten predict boxes together to form a single Human Intelligence Task (HIT). The auto-approval mechanism needs ground-truth labels to be hidden in the last task. These groundtruth labels are obtained from manual labeling which comes from the initial annotations. By comparing the worker's bounding box to the ground-truth bounding box, we implement an auto-approval process. We compute the intersection over union (IOU) of the two bounding boxes and accept the worker's annotations only if the IOU score is greater than a threshold of 0.8. LSUN \cite{LSUN_Construction_of_a_Large_scale_Image_Dataset_using_Deep_Learning_with_Humans_in_the_Loop}, proposes a similar method, using the hidden ground-truth to review the MTurk labeling results although they use the entire image as a labeling task, while we use every single object.

\subsubsection{Double-check identifications}
The incorrect classification of objects can lead to incorrect training, so even if a sub-task has passed the hidden ground-truth test, we would still double-check the class it chooses. If the selected class is different from the predicted class, add the sub-task to the republish list. Meanwhile, the class of the prediction box is changed to the one selected by the current worker. This means that the class of the object is determined only if two consecutive workers choose the same category. Otherwise, the prediction box would be repeatedly republished under this mechanism. If an object is actually a background, it would be republished at least twice to fully determine that it is the background.

\subsection{Training on half-labeled images}
In the first iteration where the prediction is based on the initial model, all the object instances in the images will not be discovered, and the accuracy of the predictions cannot be guaranteed. This is because the initial model is trained only on a small seed dataset which is insufficient to completely train the model. These predictions are sent to MTurk for correction. The new bounding boxes are used to supervise the training of our deep learning model which in turn makes new and more accurate predictions. Since the object labels of the images are incomplete there are some issues that arise when we cannot carry out the training process as it is. We make modifications to the training phase in terms of feeding appropriate training data and loss functions to suit our iterative labeling process as follows.

\subsubsection{Avoiding negative mining of potential objects}

During the training process of a object detection model, if an object is not labeled in the images, it would be treated as a background class implicitly. This is especially true for algorithms such as SSD\cite{} and Faster R-CNN \cite{Faster_R_CNN_Towards_Real_Time_Object_Detection_with_Region_Proposal_Networks} which use negative hard sampling for training the background class. In SSD, the top N highest confidence predictions that do not match any ground-truth are chosen and trained as negative samples. Faster R-CNN would randomly select a certain percentage of prediction boxes without matching ground-truths as negative samples. This would cause serious issues with training because if half of the objects are not labeled in the image, it will cause the trained model to not converge due to the wrong labels.

\begin{figure}[!t]
\centering
\includegraphics[width=3.2in]{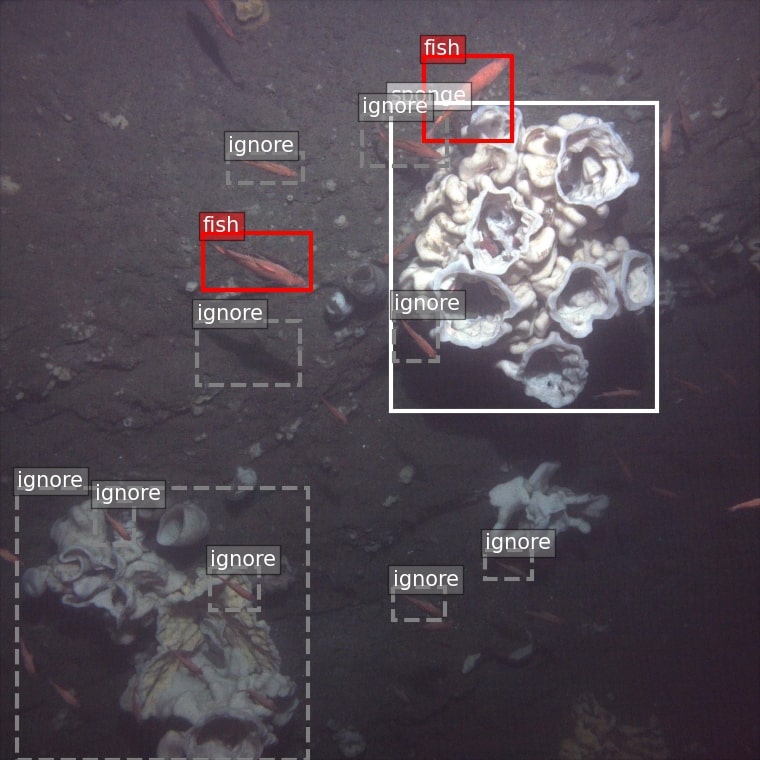}
\caption{Ignoring potential objects: The dashed bounding boxes represent the potential objects which have a significant classification score, but have not been verified and labeled yet. These unlabeled objects will be considered as negative samples implicitly during the training iterations resulting in an incorrect model. We detect these potential objects based on a threshold on their classification score and omit them from the training process. }
\label{ignore_potential_objects}
\end{figure}

The solution is to identify unlabeled potential objects and avoid training them as negative samples. If the prediction confidence score of an anchor exceeds a specific threshold and no ground-truth can match that prediction, it means that the model thinks there is a potential object here. So, it should be ignored in the training process to be discovered later as shown in \cref{ignore_potential_objects}. In the Region Proposal Network (RPN) of Faster R-CNN, we marked all the prior anchors whose confidence score exceeds 0.9 without the ground-truth label as "ignored". We ignored them in selecting negative samples for loss calculations. We also prevent them from being selected to enter the next stage of the region of interest (ROI) layer.

\begin{figure*}[!t]
\centering
\includegraphics[width=6in]{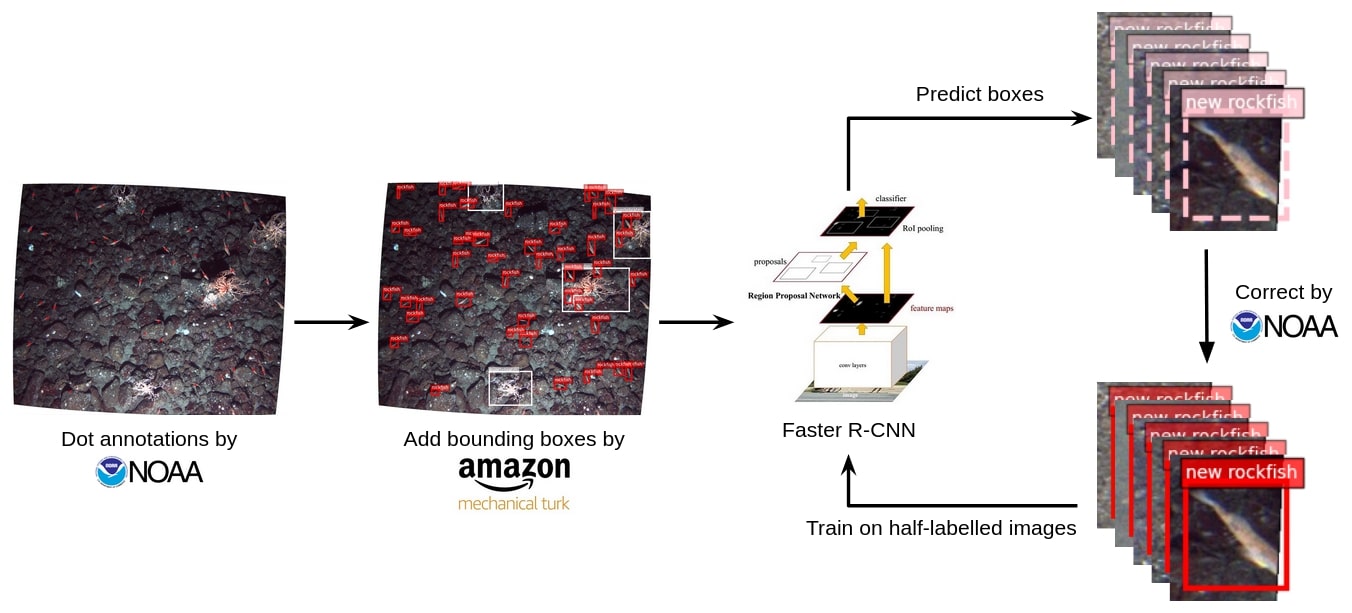}
\caption{Iterated labeling can also handle historical datasets. We can modify the original algorithm for datasets that were expert annotated with dot annotations in the past. We simply extend the bounding boxes for dot annotations and as the dataset is almost complete, we can train the half-labeled images directly, without an initial model. Objects can be predicted with this model to obtain new detections. The original NOAA annotators evaluate and correct these new detections. After updating these new labels, if there are still unlabeled objects, we can train these new labels and start a new loop.}
\label{diagram_2}
\end{figure*}

\subsubsection{Training background labels}
In the previous section we described how to avoid training potential true positive predictions as a background class. In this section we discuss how to correctly sample the background class. During the iterative labeling process some predictions would be false positives but corrected as "background" by the MTurk workers. These background labels can be used in the training process. In the Region Proposal Network (RPN), instead of randomly selecting negative samples, the boxes updated as the "background" class from MTurk auto-approval process should be trained. When the number of negative samples is significant, the probability of being trained as a potential object is very low. This is valuable because it increase the precision of our object detection model as well as avoids ignoring potential objects. We do not calculate the localization loss of background labels as they are negative samples,and their use ends with the RPN.

\subsubsection{Data augmentation}
We also perform data augmentation to generate more training samples. All the images are put through the following transformations: a flip of the image horizontally and vertically, adjustments to brightness by scaling the intensity randomly between 0.8~1.2 and a random crop of 0.8~1 of the image size.

\subsubsection{Loss Function}
Taking into account the above mentioned changes to the training phase, the loss function can be divided into four components:
\begin{itemize}
    \item The classification loss, $\sum_{i}L_{cls}(p_{i},p_{i}^*)$, where the predicted labels have object class ground truths associated with them. This Ground truth bounding boxes are obtained from MTurk after auto-approval.$N$: RPN mini-batch size
    
    \item The classification loss $\sum_{j}L_{cls}(p_{j},p_{j}^*)$, where a background class ground truth box is associated with the predicted label. This Ground truth is also obtained from MTurk after the auto-approved label is selected as back-ground.
    
    \item The classification loss $\sum_{k}L_{cls}(p_{k},p_{k}^*)]$, where the predicted box does not have any ground truth box associated with it but the prediction score with respect to an object class is lesser than 0.9. In this case we consider this as a negative sample.
    
    \item The regression loss $\lambda\frac{1}{N}\sum_{i}L_{reg}(t_{i},t_{i}^*)$ which is computed for the predicted labels which have object class groundtruth boxes associated with them.  
\end{itemize} 

Putting all the components together the loss function can be written as 
\[
\begin{aligned}
L(\{p_{i}\},\{p_{j}\},\{p_{k}\},\{t_{i}\})=&\frac{1}{N}[\sum_{i}L_{cls}(p_{i},p_{i}^*)+\\
&\sum_{j}L_{cls}(p_{j},p_{j}^*)+\\
&\sum_{k}L_{cls}(p_{k},p_{k}^*)]+\\
&\lambda\frac{1}{N}\sum_{i}L_{reg}(t_{i},t_{i}^*)
\end{aligned}
\]

The classification loss is:
\[
L_{cls}=-[p^*\cdot log(p)+(1-p^*)\cdot log(1-p)]
\]

The localization loss is:
\[
L_{reg}=
\begin{cases}
0.5|t-t^*|^2,&if\ |t-t^*|<1\\
|t-t^*|-0.5,&otherwise
\end{cases}
\]
   
where 
$i$ is the index of an anchor in a mini-batch, whose ground-truth is an object

$j$ is the index of an anchor, whose ground-truth is a labeled background

$k$ is the index of an anchor, which has no ground-truth and $p_{k}$ is lower than the ignore threshold

$p_{i,j,k}$ is the predicted probability of being an object.

$p_{i,j,k}^*$ is the ground-truth probability where 1 indicates that it is foreground. 0 means background. Here $p_{i}^*=1, p_{j,k}^*=0$.

$t_{i}$ is a vector representing the 4 parameterized coordinates of the prediction bounding box

$t_{i}^*$ is the ground-truth box associated with a positive anchor

$\lambda$ is the balancing parameter of object and localization loss.

\begin{table*}[!t]
\renewcommand{\arraystretch}{1.3}
\caption{Comparison between iterative labeling process and professional annotation platform. Our method shows superior performance over the professional annotation service. This is true not only for the quantity of labels, but also their quality as illustrated by mAP of the model prediction. }
\label{iterative_labeling_results}
\centering
\begin{tabular}{|c|c|c|c|c|c|}
\multicolumn{6}{c}{Our approach}\\
\hline
Loop & Rockfish & Starfish & Sponge & Background & mAP/50\\
\hline
1 & 2235(+2235) & 1470(+1470) & 2646(+2646) & 451(+451) & 0.8179\\
2 & 2546(+311) & 1528(+58) & 2792(+146) & 548(+97) & 0.8200\\
3 & 2794(+248) & 1576(+48) & 2825(+33) & 633(+85) & 0.8263\\
4 & 2979(+185) & 1588(+58) & 3034(+209) & 804(171) & 0.8233\\
5 & 3128(+149) & 1599(+12) & 3173(+139) & 916(+112) & 0.8410\\
6 & 3188(+60) & 1604(+5) & 3346(+173) & 969(+53) & 0.8458\\
\hline
\multicolumn{6}{c}{Pro Annotation Service}\\
\hline
& Rockfish & Starfish & Sponge & Background & mAP/50\\
\hline
& 2713 & 1600 & 1713 & / & 0.7807\\
\hline
\end{tabular}
\end{table*}


\section{Results}

\subsection{Labeling a new image dataset}
\begin{figure}[!t]
\centering
\subfloat[]{\includegraphics[width=0.8in]{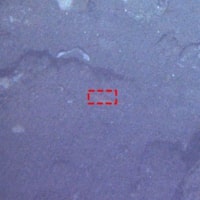}}
\hfil
\subfloat[]{\includegraphics[width=0.8in]{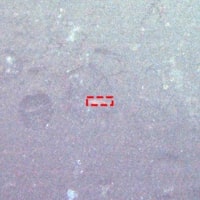}}
\hfil
\subfloat[]{\includegraphics[width=0.8in]{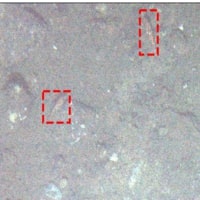}}
\hfil
\subfloat[]{\includegraphics[width=0.8in]{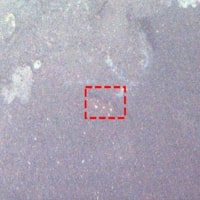}}
\caption{New detection examples. Usually these are very small fish and/or close to the background color.}
\label{new_detections}
\end{figure}

We label the 2,026 empty images with the iterative labeling process with the results as shown in Table \ref{iterative_labeling_results}. As the loop iterations increase, more and more labels are created. We also sent the same image set to the professional annotation platform to label for comparison. As we can see, the quantity of labels generated by iterative labeling process is much larger than the number of professional annotations. For the starfish that are easier to identify, the number of labels is almost the same. But for sponges that are difficult to identify and small fish close to the background color (\cref{new_detections}), the iterative labeling process successfully marked them, whereas the professional annotation platform did not. In addition, the professional annotation platform does not label false positives as background.

\begin{figure}[!t]
\centering
\subfloat[]{\includegraphics[width=0.8in]{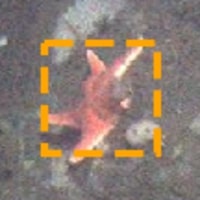}}
\hfil
\subfloat[]{\includegraphics[width=0.8in]{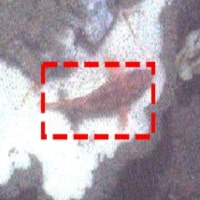}}
\hfil
\subfloat[]{\includegraphics[width=0.8in]{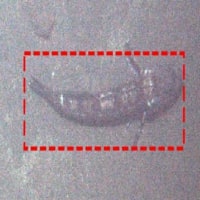}}
\hfil
\subfloat[]{\includegraphics[width=0.8in]{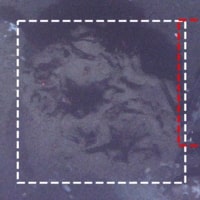}}
\caption{Missed detection examples. These are all examples of species that do not exist in the initial data set.}
\label{missed_detections}
\end{figure}

We also train the labels generated in each loop, and test the model's mAP (mean Average Precision), as shown in the last column of the Table. \ref{iterative_labeling_results}. We use the whole initial dataset to be the test set because we don't have a "ground-truth" for the empty dataset. The mAP is growing gradually, proving that the more objects we label, the better the model. Note that the mAP of the model trained from the professional labels is always lower than the model trained from the iterative labeling process.

\begin{figure*}[!t]
\centering
\includegraphics[width=7in]{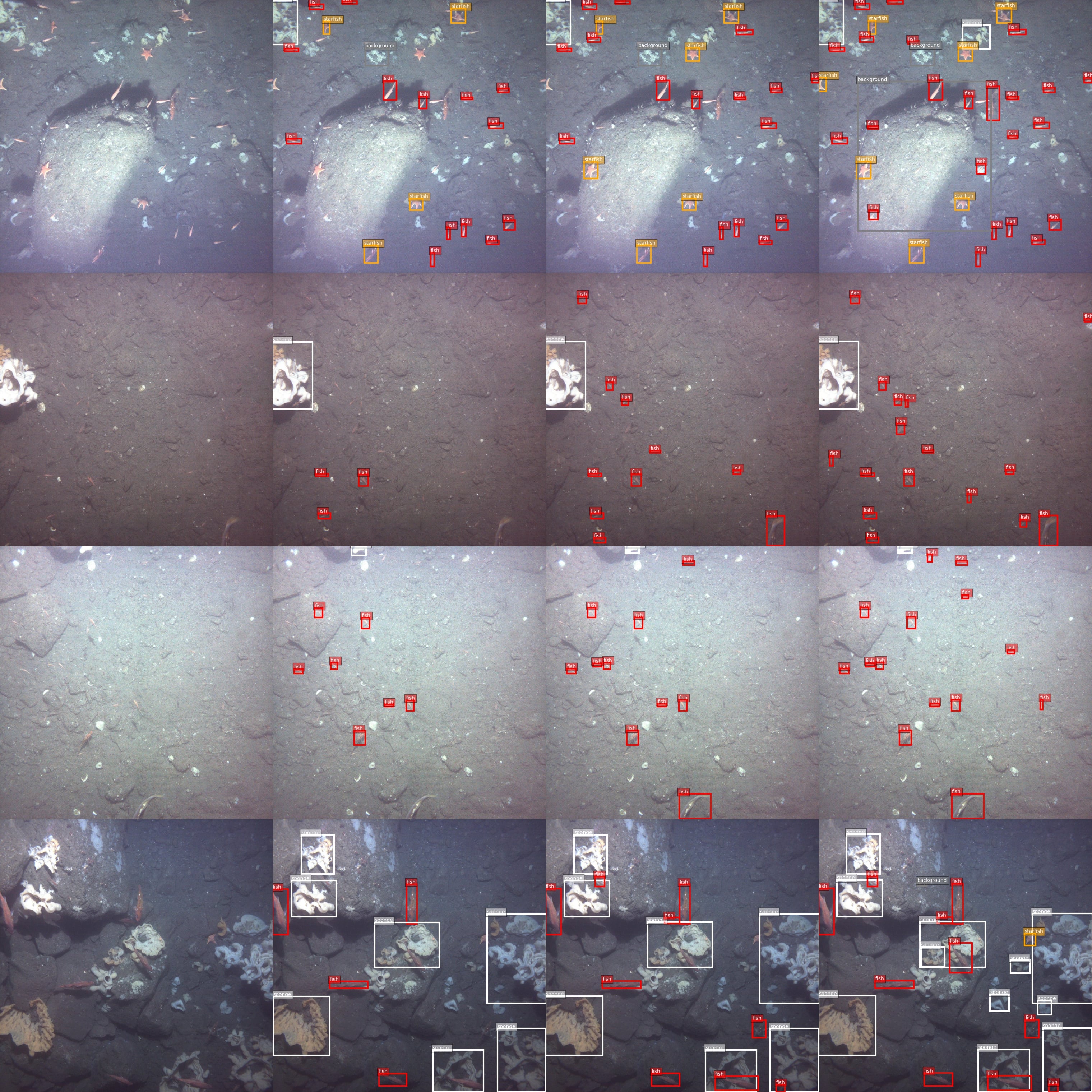}
\caption{Examples of running iterative labeling process on empty image set.The columns show individual images initially and how they are labeled after one, two and three loop iterations.}
\label{examples}
\end{figure*}

Fig. \ref{examples} shows some examples of the iterative labeling process on aseries of images. At first only a few objects are labeled. Eventually, though, with the help of our approach, all the objects in the images are labeled, resulting in a complete and well labeled data set.

\begin{figure}[!t]
\centering
\subfloat[Dot annotations given by NOAA biologists]
{\includegraphics[width=3.2in]{NOAA_dot_annotation}\label{NOAA_dot_annotation_2}}
\hfil
\subfloat[Extend bounding box with MTurk, detect missed rockfish with the Iterative labeling Process and complete the dataset.]
{\includegraphics[width=3.2in]{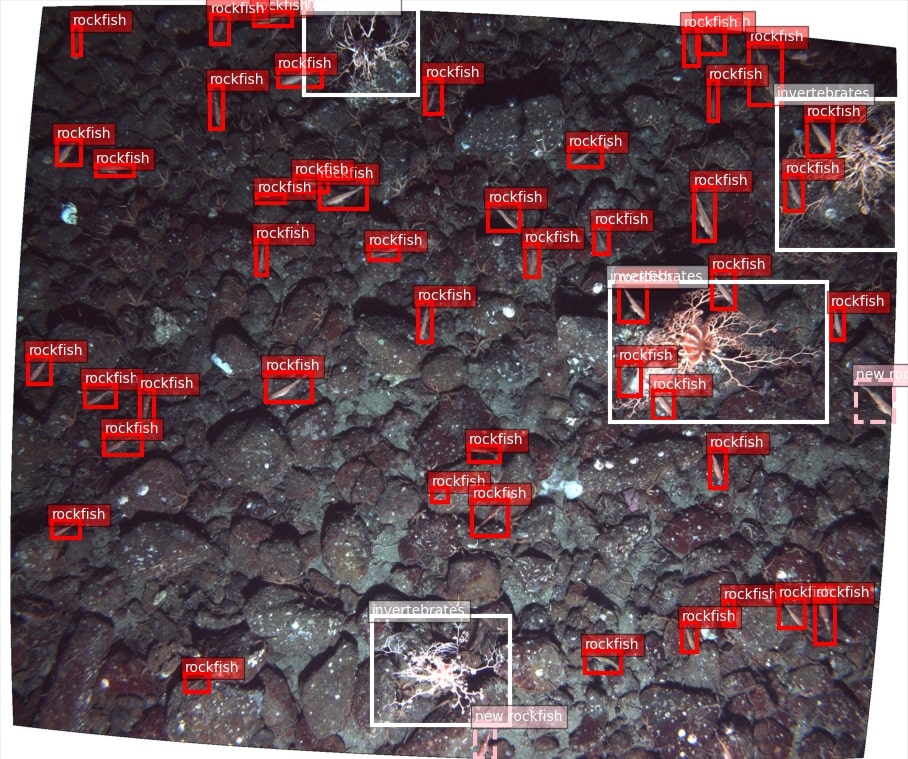}\label{NOAA_predict}}
\caption{The results of the iterative labeling process on a single image}
\end{figure}

\begin{figure}[!t]
\centering
\subfloat[]{\includegraphics[width=0.8in]{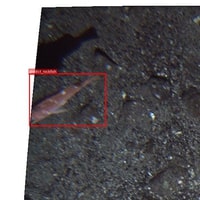}}
\hfil
\subfloat[]{\includegraphics[width=0.8in]{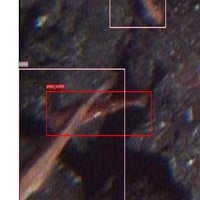}}
\hfil
\subfloat[]{\includegraphics[width=0.8in]{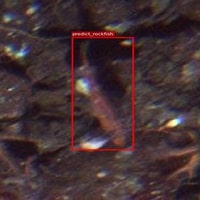}}
\hfil
\subfloat[]{\includegraphics[width=0.8in]{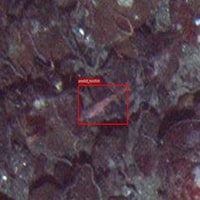}}
\caption{Missed Rockfish examples fall in three classes of roughly equal numbers. The first class as illustrated in (a) and (b) consists of those at the border of an image with the head missing. The second class illustrated in (d) corresponds to a similar number of small fish that are hard to detect while the third class encompassed missed detections that did not fall into the first two classes as shown in (c). We note that the headless fish corresponding to the first class were deliberately not labeled by the expert annotators in the NOAA dataset.}
\label{missed_rockfish}
\end{figure}

\begin{figure*}[!t]
\centering
\includegraphics[width=7in]{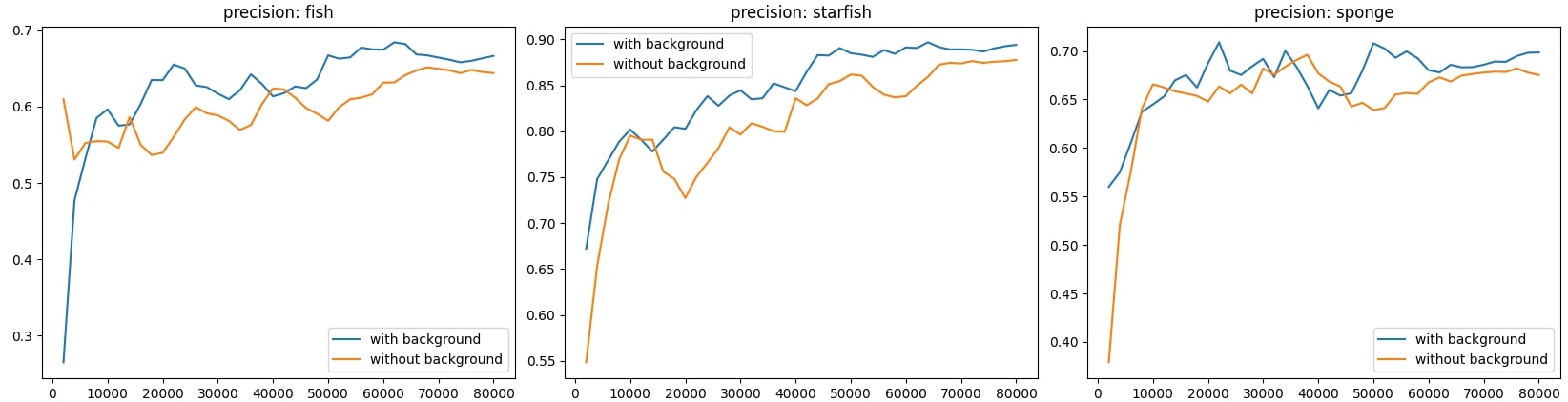}
\caption{Precision of fish, starfish and sponge. The blue lines are the model precision if we train the background labels. The orange lines are the model precision trained without background labels. They prove that train background labels could reduce the false positives and improve the accuracy.}
\label{precision}
\end{figure*}

However, there are some limitations to the iterative labeling process. Some objects labeled by the professional annotation platform have not been detected in the iterative labeling process. Fig. \ref{missed_detections} is a representation of these objects that are not covered. The first starfish has a different shape from other tiled starfish. The second one is a fish swimming above the white sponge. Due to the lack of fish in this particular background in the initial training set, almost all such fish on a white sponge were not detected. This is the case where most fish are not labeled. The third picture is a different fish species, which does not appear in the initial dataset either. The fourth object is a dead sponge which is typically not labeled by the marine community. 

\subsection{A legacy dot annotated dataset}

We also have a dataset with dot annotations provided by NOAA marine biologists (\cref{NOAA_dot_annotation_2}). In this case, we simplify the process as shown in \cref{diagram_2}. First, we publish these dot labels to MTurk workers using our assistive interface (\cref{assistive_interface}). The workers give us tight and accurate bounding boxes which are available for deep learning. Since the labels in this dataset are almost complete, we train the half-labeled images directly, without an initial model. We predict objects with this model and get new detections. NOAA scientists then checked these new detections and update these new labels to complement the missed objects (\cref{NOAA_predict}).

The detailed numbers for the NOAA dataset are shown in Table \ref{NOAA_dataset}. We ran the iterative labeling process on the rockfish class to find missing objects. Although it is a fairly complete dataset, we could still find 1644 missing rockfish, as shown in \cref{missed_rockfish}. We note that a significant fraction of these "missing" rockfish (721 or 0.8\% of the total) actually correspond to the case of rockfish without a head which are purposely not labeled by marine biologists as they rely on the fish's head to make distinctions at the species level. A smaller number (521 or 0.6\% of the total) of the missing rockfish were either similar in color to the background and or very small. A further 149 belonged to both categories, while 549 (0.6\% of the total) were reasonably sized, complete but failed to be detected.   

We should also point out that the NOAA dataset was annotated to a greater level of taxonomic resolution than we present here. Classification of the data to such levels is an interesting and open problem beyond the scope of this work. 

\begin{table}[!t]
\renewcommand{\arraystretch}{1.3}
\caption{Labeling results for the NOAA dataset. (+1644) refers to the additional 1644 Rockfish that were labeled with the iterative labeling process.}
\label{NOAA_dataset}
\centering
\begin{tabular}{|c|c|}
\hline
Class & Labels\\
\hline
Sponges & 101399\\
Rockfish & 91432(+1644)\\
Corals & 40180\\
Flatfish & 4688\\
Roundfish & 3828\\
Other Invertebrates & 150\\
Skates & 10\\
\hline
\end{tabular}
\end{table}


\subsection{Train background labels}
The main purpose of training background labels is to reduce false positives, in other words, increase the precision of the model. We train the dataset with background labels and without them to compare the precision of the model. The comparisons of precision of each category are shown in Fig. \ref{precision}. The precision of all the categories has been improved. 

\section{Conclusion}
In this paper we have presented a method for quickly labeling large underwater datasets and show that this method is robust, precise and reasonable in terms of overall effort for a large number of images and classes. We started with an empty image set and showed how the iterative labeling process generates bounding box annotations gradually to eventually return a complete dataset with rockfish, starfish and sponge annotations with a small number of iterations. 

We also took a NOAA dataset which only had dot annotations on it. We utilized MTurk workers to extend the dots to bounding boxes with the help of an assistive labeling interface. We could then run iterative labeling process to detect rockfish in the dataset that had not been labeled. These missing objects were validated by NOAA researchers to complete the dataset.

Both datasets are freely available for other researchers to use via the website \cite{fieldrobotics_datasets} and can serve as a benchmark for validating different machine learning methodologies for fisheries related applications underwater. 


\ifCLASSOPTIONcaptionsoff
  \newpage
\fi

\bibliographystyle{IEEEtran}
\newpage
\bibliography{references}

\end{document}